\documentclass[a4paper,onecolumn]{IEEEtran}

\usepackage{amsmath,amssymb,amsfonts}
\usepackage{mathrsfs}
\usepackage{graphicx}
\usepackage{tabularx}
\usepackage{hyperref}
\usepackage{subcaption}
\usepackage{textcomp}
\usepackage{smartdiagram}
\usepackage{adjustbox}
\usepackage{pgfplots}
\usepackage{float}
\usepackage{xcolor}
\usepackage{bm}
\usepackage[boxed,commentsnumbered, linesnumbered,ruled,vlined]{algorithm2e}
\usepackage[style=ieee, backend=biber, natbib=true, maxbibnames=1, minbibnames=1]{biblatex}
\usepackage{siunitx}
\usepackage{tikz}
\usepackage{circuitikz}
\usepackage{dirtytalk}
\usepackage[acronym]{glossaries} 
\usepackage{svg}
\usepackage{listings}
\usepackage{setspace}
\usetikzlibrary{shapes.geometric, arrows, positioning}
\usepackage{titlesec}
\titlespacing{\subsubsection}{0pt}{10pt}{5pt}

\usepackage[normalem]{ulem}

\usepackage[left=2.2cm, right=2.2cm, top=3.5cm, bottom=2.5cm]{geometry}

\usepackage[skip=5pt plus1pt, indent=10pt]{parskip}

\usepackage{fancyhdr}

\fancypagestyle{firstpage}{%
	\fancyhf{}%
	\fancyhead[C]{}
	
	\fancyfoot[C]{}
}

\graphicspath{{./figures/}}

\definecolor{ForestGreen}{RGB}{34,150,34}
\newacronym{SoC}{SoC}{System-on-Chip}
\newacronym{SoCs}{SoCs}{System-on-Chips}
\newacronym{IPs}{IPs}{Intellectual properties}
\newacronym{FPV}{FPV}{Formal Property Verification}
\newacronym{CSR}{CSR}{Control/Status Register}
\newacronym{IP}{IP}{Intellectual Property}
\newacronym{IC}{IC}{Integrated Circuit}
\newacronym{UML}{UML}{Unified Modelling Language}
\newacronym{NRE}{NRE}{Non-Recurring Engineering} 
\newacronym{AMS}{AMS}{Analog Mixed-Signal}
\newacronym{SPI}{SPI}{Serial Peripheral Interface}
\newacronym{XML}{XML}{Extensible Markup Language}
\newacronym{MDA}{MDA}{Model Driven Architecture}
\newacronym{CIM}{CIM}{Computation Independent Model}
\newacronym{APIs}{APIs}{Application Programming Interface}
\newacronym{PSM}{PSM}{Platform Specific Model}
\newacronym{PIM}{PIM}{Platform Independent Model}
\newacronym{SECDED}{SECDED}{Single Error Correction Double Error Detection}
\newacronym{DECTED}{DECTED}{Double Error Correction Triple Error Detection}
\newacronym{SVA}{SVA}{System Verilog Assertion}
\newacronym{EDA}{EDA}{Electronic Design Automation}
\newacronym{TCL}{TCL}{Transaction Control Language}
\newacronym{SST}{SST}{State Space Tunneling}
\newacronym{ABVIP}{ABVIP}{Assertion Baed Verification IP}
\newacronym{JG}{JG}{Jasper Gold}
\begin{document}

\title{Analogous Alignments:  Digital \say{Formally} meets Analog\\}

\author{
\IEEEauthorblockA{\vspace{4mm} Hansa Mohanty,
Infineon Technologies,
Dresden, Germany
(\textit{hansa.mohanty@infineon.com})}\\
\IEEEauthorblockA{\vspace{4mm} Deepak Narayan Gadde,
Infineon Technologies,
Dresden, Germany
(\textit{deepak.gadde@infineon.com})}}

\maketitle

\thispagestyle{fancy}

\lhead{\begin{picture}(-53,0) \put(-53,0){\includegraphics[height=2.2cm]{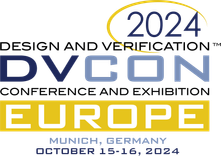}} \end{picture}}

\begin{abstract}
\textbf{\emph{Abstract}\!
\textemdash  The complexity of modern-day \acrfull{SoCs} is continually increasing, and it becomes increasingly challenging to deliver dependable and credible chips in a short time-to-market. Especially, in the case of test chips, where the aim is to study the feasibility of the design, time is a crucial factor. Pre-silicon functional verification is one of the main contributors that makes up a large portion of the product development cycle \cite{VerStudy}. Verification engineers often loosely verify test chips that turn out to be non-functional on the silicon, ultimately resulting in expensive re-spins. To left-shift the verification efforts, formal verification is a powerful methodology that aims to exhaustively verify designs, giving better confidence in the overall quality. This paper focuses on the pragmatic formal verification of a mixed signal \acrfull{IP} that has a combination of digital and analog blocks. This paper discusses a novel approach of including the analog behavioral model into the formal verification setup. Digital and \acrfull{AMS} designs, which are fundamentally different in nature, are integrated seamlessly in a formal verification setup, a concept that can be referred to as \say{Analogous Alignments}. Our formal setup leverages powerful formal techniques such as \acrfull{FPV}, \acrfull{CSR} verification, and connectivity checks. The properties used for \acrshort{FPV} are auto-generated using a metamodeling framework \cite{metamodel}. The paper also discusses the challenges faced especially related to state-space explosion, non-compatibility of formal with \acrshort{AMS} models, and techniques to mitigate them such as k-induction. With this verification approach, we were able to exhaustively verify the design within a reasonable time and with sufficient coverage. We also reported several bugs at an early stage, making the complete design verification process iterative and effective.}
\end{abstract}


\section{Introduction}
\label{sec:intro}

Verification has become the bottleneck in product development cycles, as it takes more than \SI{60}{\percent} of the overall project time \cite{VerStudy}. In contemporary electronic systems, analog circuits are integral components. The engineering community faces increasing challenges as the time-to-market shrinks amid the increasing complexity of mixed-signal designs. A critical strategy to optimize the verification process is simulation, employing a hierarchy of circuit models that vary in abstraction. The introduction of a behavioral model for \acrshort{AMS} simulation represents a notable abstraction method. While this model is advantageous for simulation-based verification, particularly with respect to assessing the functionality of digital components, its integration into formal verification workflow can be difficult. Formal verification employs a mathematical approach to ascertain the correctness of a design. Traditionally, this verification requires the specifier to possess an understanding of property formulation in system verilog assertions. The manual implementation of these properties leads to certain challenges. Not only is it prone to human error, but it can also be monotonous and laborious, especially when dealing with extensive state spaces. To mitigate these issues, properties have been auto-generated utilizing a Metamodeling framework. Metamodeling serves as an automation framework designed to reduce \acrfull{NRE} costs. This framework leverages \acrfull{UML} diagrams, Python, and Mako templates to create diverse outputs across various programming languages. This paper delves into the details of automating property generation. Formal verification plays a crucial role in validating hardware circuits and in revealing numerous minor yet significant features, such as x-propagation, register, and connectivity issues. However, its effectiveness is diminished when applied to verifying \acrshort{AMS} circuits. In the context of digital verification, the analog components primarily function to deliver control and status signals, which collectively ensure the correct operation of the \acrshort{IC}. During digital simulations, the performance of the analog circuitry is not the focal point. In this paper, we provide insights into two methods firstly assuming the ideal behavior of the \acrshort{AMS} behavioral model and the secondly a novel method that incorporates a formal-friendly analog model, which is conducive to the smooth integration into formal verification frameworks.

\section{Design Overview}
\label{sec:do}
\begin{figure}[h!]
\centering
  \includegraphics [width=0.4\textwidth] {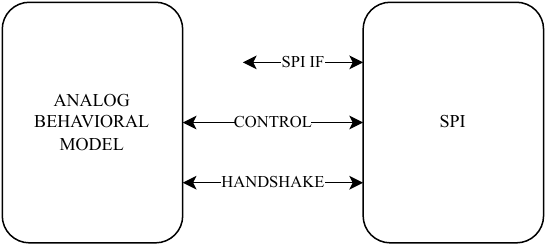}
\caption{Abstract level overview of the design}
\label{design_block_dig}
\end{figure}
The \acrshort{IP} consists of a \acrfull{SPI} that communicates with an \acrshort{AMS} model as shown in Fig.~\ref{design_block_dig}. In practical applications, the \acrshort{AMS} circuit is composed of a memory unit engineered to permit a single write operation followed by the capability for multiple read accesses. The constructed analog behavioral model, designated for digital simulation purposes, plays a crucial role in the transmission of specific control and handshake signals. These signals include a 'ready' indicator, which signifies that the analog circuitry is prepared to initiate the subsequent transaction and a 'busy' signal, which is asserted during the course of an active transaction. Furthermore, the \acrfull{SPI}, which incorporates a register bank, serves as a conduit for serial inputs. These inputs are then channeled to the analog behavioral model. Concurrently, the values corresponding to various status signals, which originate from the \acrshort{AMS} behavioral model, are accessible through a designated register within the register bank. This systematic approach ensures a seamless interface between the digital and analog components of the \acrshort{AMS} system, facilitating efficient communication and operational synchronization. The principal objective of the \acrfull{IP} under examination is to archive values received from various components of the \acrfull{SoC}. The iteration of the design presented here was realized as a test chip, with the primary intent of evaluating potential reductions in the area occupied by the analog circuitry. A fully equipped functional \acrshort{IP} is implemented for the tasks of data rectification and autonomous data integrity checks. This comprehensive functional \acrshort{IP} is equipped with both  \acrfull{SECDED} and \acrfull{DECTED}. It should be noted that the complete formal verification of this robust IP infrastructure is anticipated as a subject of future investigation and is beyond the scope of the current paper.

\section{Metamodel Framework}
\label{sec:meta}
A Metamodeling framework is an automation framework where a model is used to represent a system in a certain abstraction level. A metamodel is utilized to define the structure of a model and the relationships between its constituents. Metamodels extend beyond models and are used to represent models of models. Metamodeling serves as an automation framework designed to decrease  \acrfull{NRE} costs. This framework utilizes \acrfull{UML} diagrams, Python, and Mako templates to generate outputs in diverse programming languages. Metamodeling is rooted in the principles of \acrfull{MDA}, which advocates for the use of modeling techniques to enhance productivity and abstraction levels during the development process. The different abstraction levels used in \acrshort{MDA} are:  \acrfull{CIM}, \acrfull{PIM}, and \acrfull{PSM}. In the first step the specification is added to the framework. The specification is generally added in the form of a \acrshort{UML} diagram. The framework parser goes through the \acrshort{UML} diagram and generates a Python-based \acrfull{APIs}. The \acrshort{APIs} have been generated automatically, including several methods, such as setters(), getters(), and other frequently used methods, making it simpler to access the metamodel.  The
 design definition is then entered using the Metamodel Graphical User Interface (GUI), which is also generated as part of the Python and Mako templates to framework after the APIs have been created. Data generated by
 the Metamodeling GUI is usually in an \acrfull{XML} format. By using the API methods, the reader reads the specifications from the XML file, and the writer can produce viewcode in a language of their choice \cite{metamodel}. This paper presents a application of the Metamodeling framework to facilitate the generation of \acrshort{SVA}, a critical component in the verification of hardware designs. Our approach begins with the construction of a \acrshort{UML} class diagram, meticulously detailing the interconnections and relationships amongst system components. This abstraction level encapsulates a high-level view of the system, integrating essential elements such as registers, bitfields, access policies, and reset values. Subsequently, we transmute this CIM representation into an \acrshort{XML} file, affording us a versatile and universally compatible format that seamlessly feeds into the successive layer of abstraction. Advancing to the \acrshort{PIM}, the register and other specifications are  converted into an \acrshort{XML} file. Consequently advancing to \acrshort{PSM} level of abstraction, the \acrshort{XML} file is converted to a PIM.xml, file and we define a robust 'property class' which is otherwise known as a MAKO template, laying the groundwork for the automated generation of three crucial types of SystemVerilog assertions: 

\begin{itemize}
\item Register read-write operations,
functionality check between the \acrshort{SPI} and register bank,
\item Daisy chain mechanisms within the circuitry.
\item Communication protocols between the \acrshort{AMS} circuits and the \acrshort{SPI}
\end{itemize}
Each category of assertion plays a pivotal role in ensuring the integrity and correctness of digital and analog interactions. Our methodology not only expedites the verification process but also enhances its accuracy, thereby contributing to the reliability and efficiency of hardware design verification.

\begin{figure}[h!]
\centering
  \includegraphics [width=\textwidth] {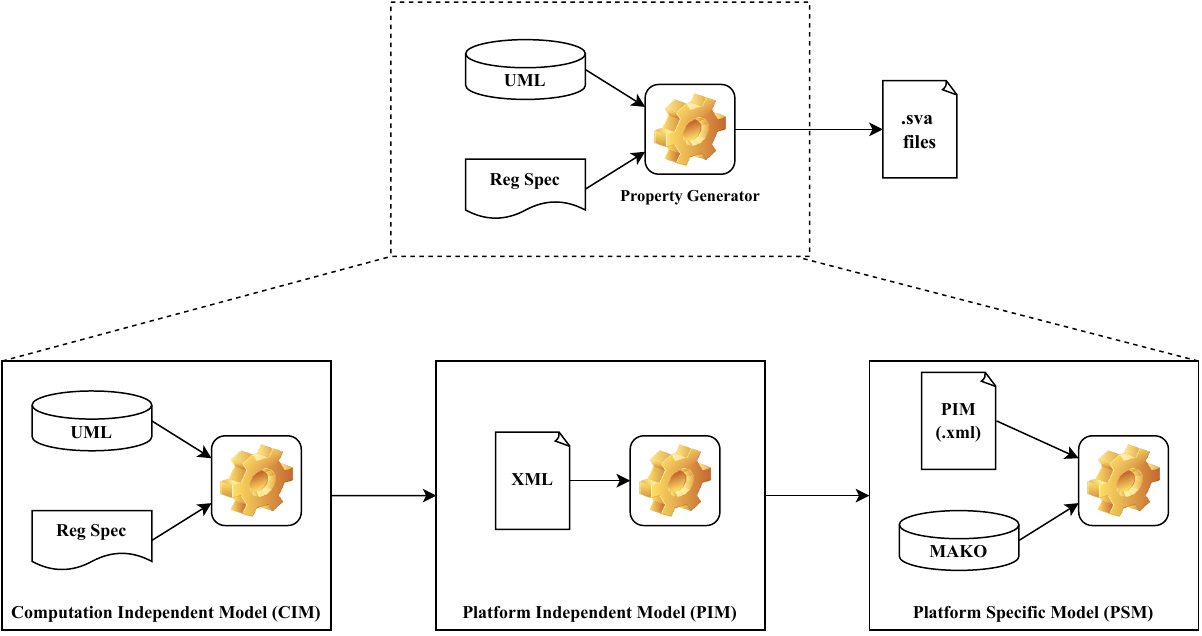}
\caption{Metamodel framework for SVA property generation}
\label{Metagen_flow}
\end{figure}
\section{Design Verification}
\label{sec:dv}

The \acrshort{IP} described in this paper is utilized  within the realm of automotive applications. The digital components of these \acrshort{IPs}, as delineated in Section \ref{sec:do}, present themselves as prime candidates for leveraging Formal Verification methodologies. Nonetheless, the incorporation of \acrshort{AMS} components presents a formidable challenge to the seamless adoption of these verification techniques into our established workflow. The core of the challenge lies in the inherent complexity and non-synthesizability of \acrshort{AMS} circuitry, which resists straightforward formal analysis. In light of this, our experiment bifurcates into two distinct investigative pathways. The initial pathway operates under the assumption that the analog components behave in an idealized manner—an assumption that simplifies the verification process but may not hold up against the details and anomalies of real-world applications. In this pathway, we also incorporate the use of additional Jasper applications during the bring-up phase, broadening the horizon of formal verification beyond its traditional routes. These applications, which extend to functions like register checks and connectivity validations, are not only user-friendly but also effective in early bug detection and ease of debugging. The second investigative pathway, which is more forward-looking, endeavors to assimilate the analog behavioral model, complete with its non-ideal characteristics, into the formal verification framework. This approach acknowledges the limitations of the ideal behavior assumption, particularly as the complexity of \acrshort{AMS} circuits escalates. Subsequent sub-sections will expound upon the methodologies and findings associated with each investigative pathway, providing a comprehensive analysis and comparison of their respective efficacies and applicabilities in the context of automotive IP verification. Through this exploration, we aim to illuminate the path towards more robust and comprehensive formal verification strategies that can accommodate the nuanced behaviors of \acrshort{AMS} circuitry. Both approaches are described in the following sections.

\subsection{Formal Verification}
Formal verification uses technologies that mathematically analyze the space of possible behaviors of a design rather than computing results for particular values \cite{fvbook}. It is an exhaustive verification technique that uses mathematical proof methods to verify whether the design implementation matches design specifications \cite{aman_dvcon_ecc}.

\begin{figure}[h!]
\centering
  \includegraphics [width=0.5\textwidth] {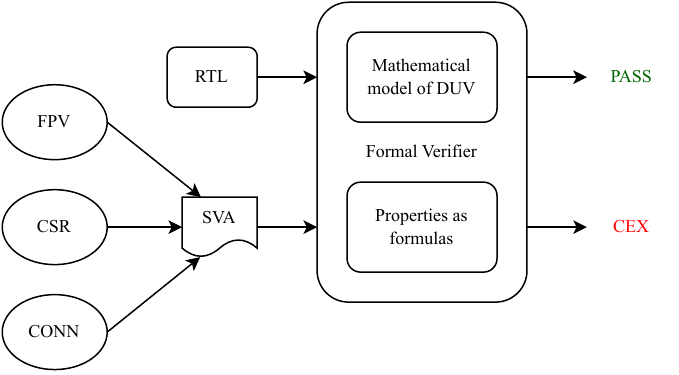}
\caption{Formal verification}
\label{fv_flow}
\end{figure}

Fig.~\ref{fv_flow} shows the working of a formal verifier. There are two inputs to the formal verifier tool. On the one hand, the Design Under Verification (DUV) is fed into the tool which is converted into a mathematical model. On the other hand, properties, written in SystemVerilog Assertions (SVA) that capture the intent of the design are fed into the tool. The tool then converts these properties into mathematical formulas. In the next step, the tool tries to prove these mathematical formulas on the mathematical model of the DUV. If the properties do not hold, the tool announces, and a Counter Example (CEX) is generated by the tool to further debug. In general, the absence of a CEX is nothing but a pass or proven result \cite{aman_dvcon_configvermet}. To meet different verification requirements, we used different formal verification techniques and apps from Cadence Jasper as shown in Fig \ref{fv_flow}. The following sub-sections provide an overview of the techniques used.

\subsubsection{Control/Status Register Verification}
The CSR verification app from Cadence Jasper offers a pragmatic solution for verifying the registers in the DUV. It is especially beneficial at the early stage of the project when the UVM testbench is not fully functional, yet we need to verify the correct behavior of the registers such as the register access policies, reset values of register fields and basic register write reads. The register description is prepared by the concept engineer and is used to generate the Comma Separated Values (CSV) file, which is an input to the formal tool. Since this approach is fully automated as a push-button solution, it also helps design engineers verify their RTL for every register change before releasing the next version of the RTL. The DUV mentioned in this paper has several registers that have varied policies therefore a preliminary check gives confidence over the design. A major benefit of using a formal based approach for register verification is the ease of debugging. The CEXs in case of a failure are usually within 10 clock cycles which helps to identify the root cause easily. However, there are also limitations associated with this approach. Usually, the properties generated by the tool are encrypted and cannot be accessed. Additionally, custom register schemes that are specific to the project cannot be verified with this approach as the properties are only generated for standard register types.

\subsubsection{Connectivity Verification}
The CONN application from Cadence Jasper is a useful tool for checking connections in circuits, specifically designed for both simple and conditional links. This tool is part of the formal verification process, known for its thoroughness, ensuring that no connection is overlooked. In our project, we paid special attention to the links between the analog parts (like sensors) and digital parts (like processors) of our design. We made sure that these connections were properly established from start to finish. The connectivity checker within the Jasper suite is a program that helps us do this automatically. It creates a list of checks that need to be made based on information we provided in a table. This table is straightforward — it lists where the signal starts (the source block), the name of the signal, where the signal is going (the destination block), the name of the signal there, and any special conditions that must be met for the connection to work. Once we fill out this table, we implement a tcl script, which is a small computer program that uses the table to check everything automatically with the Jasper tool. We can also do the checks backward using a feature where the app figures out the connections and fills in the table for us. This is especially helpful to double-check our work and make sure it matches the original plans given by the concept engineer.  

\subsubsection{Formal Property Verification}
The Jasper application is a tried-and-true method for ensuring that computer systems perform their functions correctly through a formal verification process. This particular application stands out because it requires users to have a foundational understanding of \acrshort{SVA}, which are essentially rules that guide the verification process. To operate this tool, users must provide three essential components: the \acrshort{SVA} specification written in the form of properties, a control file (\acrshort{TCL} file), and the hardware design tested. The application boasts substantial adaptability, making it suitable for a range of industrial applications, regardless of their scale. An impressive feature of this Jasper tool is its ability to allow users discretion in selecting different solver engines for evaluating each specification/formula. These engines are algorithmic methods that the tool uses to ensure the system adheres to the specification provided. Moreover, formal property verification offers a strategy for simplifying the complexity of the design under verification, which can make the verification process more efficient. An advanced technique for simplification will be introduced in the following sections of the paper. Venturing further, the tool is designed to accommodate explorations into areas traditionally challenging for formal verification, such as \acrshort{AMS} behavioral models, which simulate complex circuit behavior. Initially, this seemed impractical, yet a comprehensive review revealed that adopting an \acrshort{AMS} model compatible with formal verification would be less cumbersome than creating a new simulation setup altogether. The ensuing sections of the paper will delve into the initial obstacles encountered when implementing this approach and the solutions devised to overcome them. A particular focus will be on maintaining the authenticity of the system's behavior throughout the verification process, ensuring that the simplifications do not alter the expected outcomes.

\singlespacing
\noindent\underline{\textit{\textbf{Unaligned:}}}

FPV is the most commonly used app to verify the correct behavior of the design. This is a traditional method of formal verification where the features/specifications are written in the form of properties. An overview of the types of properties written is given in \ref{sec:meta}. Although all of these properties are autogenerated the output file still had many lines of the same code, as the \acrshort{SPI} is a serial communication, and the settings (polarity and phase) made the data transmission and reception in every second clock cycle. Therefore to avoid the same lines of code written repetitively for each register (read, write, and its effect on the digital/analog interface), system verilog directive sequence was used. The example sequence is given in \ref{spi_read}.

\begin{lstlisting}[language=Verilog, float=h!, caption=\acrshort{SVA} sequence to check the actual with expected output, basicstyle=\fontsize{10}{12}\selectfont\ttfamily, label={spi_read}]
//==============================================================
// SPI read data sequence
//==============================================================
  sequence spi_read(logic [15:0] data, int cycles);//checking actual with expected
    ($past(`miso_out,30+cycles) == data[15]//adding the subsequent number of cycles 
      && $past(`miso_out,28+cycles) == data[14] //because of a dummy transaction
      && $past(`miso_out,26+cycles) == data[13]
      && $past(`miso_out,24+cycles) == data[12]
      && $past(`miso_out,22+cycles) == data[11]
      && $past(`miso_out,20+cycles) == data[10]
      && $past(`miso_out,18+cycles) == data[9]
      && $past(`miso_out,16+cycles) == data[8]
      && $past(`miso_out,14+cycles) == data[7]//Comparing the MISO output with 
      && $past(`miso_out,12+cycles) == data[6]// data written in write transaction
      && $past(`miso_out,10+cycles) == data[5]
      && $past(`miso_out,8+cycles) == data[4]
      && $past(`miso_out,6+cycles) == data[3]
      && $past(`miso_out,4+cycles) == data[2]
      && $past(`miso_out,2+cycles) == data[1]
      && $past(`miso_out,cycles) == data[0]);
  endsequence
\end{lstlisting}

Building on the mechanism of data transmission and reception, the data integrity checks required write and then read cues from the registers and then influence on the Digital-Analog interface. The operation of subsequent read after write requires an intermediate write/read operation which is a dummy. However, the register address can't be invalid as error handling is not a part of the specification. Therefore, this intermediate transaction is a valid one. The bitfield for defining the transmission to be read/write was exercised by the tool, which increased the state space to manifold. Therefore, the abstraction technique of k-induction was used.  Induction is a method to check if the design is in a random, good state and whether it will be in a good state at the next cycle \cite{cadence}. K-induction bounded model checking consists of two steps. These are the base case and the induction step. Firstly, rather than being just one state, the property is assumed to hold for a path of $n$ successive states. This means that a more extensive base-case must be demonstrated. Secondly, the path's states are assumed to be distinct. Finiteness implies that the second strengthening completes the method in the sense that there is always a length for which the induction-step is provable, this can be formalized as equations (\ref{base}) and (\ref{step}) \cite{induction_paper1}:

\begin{equation}
\text {Base}_n \quad:=\mathbf{I}_0 \wedge\left(\left(\mathbf{P}_0 \wedge \mathbf{T}_0\right) \wedge \ldots \wedge\left(\mathbf{P}_{n-1} \wedge \mathbf{T}_{n-1}\right)\right) \wedge \overline{\mathbf{P}_n}
\label{base}
\end{equation}

\begin{equation}
\text {Step}_n \quad:=\left(\left(\mathbf{P}_0 \wedge \mathbf{T}_0\right) \wedge \ldots \wedge\left(\mathbf{P}_n \wedge \mathbf{T}_n\right)\right) \wedge \overline{\mathbf{P}_{n+1}}
\label{step}
\end{equation}

where ${I}_0$ is the initial state, ${P}_0$ is the property in it's initial state  ${T}_0$ is time zero, and $n$ is the time step. The interpretation of these formulas/equations is mentioned in Fig.~\ref{formula}. If the $n$-th base-case is unsatisfiable, the statement should be interpreted as \say{There exists no $n$-step path to a state, violating the property, assuming the property the first $n$-1 steps}. If the $n$-th induction-step is unsatisfiable, it should be read as \say{Following an $n$-step trace where the property holds true, there exists no next state where it fails}. SAT solvers are used in modern \acrshort{EDA} tools to solve such induction equations.

\tikzset{every picture/.style={line width=1pt}} 
\begin{figure}[h!]
\centering
\begin{tikzpicture}[x=0.75pt,y=0.75pt,yscale=-1,xscale=1]

\draw   (132,123) .. controls (132,118.03) and (136.03,114) .. (141,114) .. controls (145.97,114) and (150,118.03) .. (150,123) .. controls (150,127.97) and (145.97,132) .. (141,132) .. controls (136.03,132) and (132,127.97) .. (132,123) -- cycle ;
\draw   (170,123) .. controls (170,118.03) and (174.03,114) .. (179,114) .. controls (183.97,114) and (188,118.03) .. (188,123) .. controls (188,127.97) and (183.97,132) .. (179,132) .. controls (174.03,132) and (170,127.97) .. (170,123) -- cycle ;
\draw    (141,114) .. controls (156.68,83.6) and (170.09,101.94) .. (177.83,112.42) ;
\draw [shift={(179,114)}, rotate = 233.13] [color={rgb, 255:red, 0; green, 0; blue, 0 }  ][line width=0.75]    (10.93,-3.29) .. controls (6.95,-1.4) and (3.31,-0.3) .. (0,0) .. controls (3.31,0.3) and (6.95,1.4) .. (10.93,3.29)   ;
\draw   (208,123) .. controls (208,118.03) and (212.03,114) .. (217,114) .. controls (221.97,114) and (226,118.03) .. (226,123) .. controls (226,127.97) and (221.97,132) .. (217,132) .. controls (212.03,132) and (208,127.97) .. (208,123) -- cycle ;
\draw    (179,114) .. controls (194.68,83.6) and (208.09,101.94) .. (215.83,112.42) ;
\draw [shift={(217,114)}, rotate = 233.13] [color={rgb, 255:red, 0; green, 0; blue, 0 }  ][line width=0.75]    (10.93,-3.29) .. controls (6.95,-1.4) and (3.31,-0.3) .. (0,0) .. controls (3.31,0.3) and (6.95,1.4) .. (10.93,3.29)   ;
\draw    (217,114) .. controls (232.68,83.6) and (246.09,101.94) .. (253.83,112.42) ;
\draw [shift={(255,114)}, rotate = 233.13] [color={rgb, 255:red, 0; green, 0; blue, 0 }  ][line width=0.75]    (10.93,-3.29) .. controls (6.95,-1.4) and (3.31,-0.3) .. (0,0) .. controls (3.31,0.3) and (6.95,1.4) .. (10.93,3.29)   ;
\draw   (320,123) .. controls (320,118.03) and (324.03,114) .. (329,114) .. controls (333.97,114) and (338,118.03) .. (338,123) .. controls (338,127.97) and (333.97,132) .. (329,132) .. controls (324.03,132) and (320,127.97) .. (320,123) -- cycle ;
\draw    (291,114) .. controls (306.68,83.6) and (320.09,101.94) .. (327.83,112.42) ;
\draw [shift={(329,114)}, rotate = 233.13] [color={rgb, 255:red, 0; green, 0; blue, 0 }  ][line width=0.75]    (10.93,-3.29) .. controls (6.95,-1.4) and (3.31,-0.3) .. (0,0) .. controls (3.31,0.3) and (6.95,1.4) .. (10.93,3.29)   ;
\draw   (358,123) .. controls (358,118.03) and (362.03,114) .. (367,114) .. controls (371.97,114) and (376,118.03) .. (376,123) .. controls (376,127.97) and (371.97,132) .. (367,132) .. controls (362.03,132) and (358,127.97) .. (358,123) -- cycle ;
\draw    (329,114) .. controls (344.68,83.6) and (358.09,101.94) .. (365.83,112.42) ;
\draw [shift={(367,114)}, rotate = 233.13] [color={rgb, 255:red, 0; green, 0; blue, 0 }  ][line width=0.75]    (10.93,-3.29) .. controls (6.95,-1.4) and (3.31,-0.3) .. (0,0) .. controls (3.31,0.3) and (6.95,1.4) .. (10.93,3.29)   ;
\draw   (132,243) .. controls (132,238.03) and (136.03,234) .. (141,234) .. controls (145.97,234) and (150,238.03) .. (150,243) .. controls (150,247.97) and (145.97,252) .. (141,252) .. controls (136.03,252) and (132,247.97) .. (132,243) -- cycle ;
\draw   (170,243) .. controls (170,238.03) and (174.03,234) .. (179,234) .. controls (183.97,234) and (188,238.03) .. (188,243) .. controls (188,247.97) and (183.97,252) .. (179,252) .. controls (174.03,252) and (170,247.97) .. (170,243) -- cycle ;
\draw    (141,234) .. controls (156.68,203.6) and (170.09,221.94) .. (177.83,232.42) ;
\draw [shift={(179,234)}, rotate = 233.13] [color={rgb, 255:red, 0; green, 0; blue, 0 }  ][line width=0.75]    (10.93,-3.29) .. controls (6.95,-1.4) and (3.31,-0.3) .. (0,0) .. controls (3.31,0.3) and (6.95,1.4) .. (10.93,3.29)   ;
\draw   (208,243) .. controls (208,238.03) and (212.03,234) .. (217,234) .. controls (221.97,234) and (226,238.03) .. (226,243) .. controls (226,247.97) and (221.97,252) .. (217,252) .. controls (212.03,252) and (208,247.97) .. (208,243) -- cycle ;
\draw    (179,234) .. controls (194.68,203.6) and (208.09,221.94) .. (215.83,232.42) ;
\draw [shift={(217,234)}, rotate = 233.13] [color={rgb, 255:red, 0; green, 0; blue, 0 }  ][line width=0.75]    (10.93,-3.29) .. controls (6.95,-1.4) and (3.31,-0.3) .. (0,0) .. controls (3.31,0.3) and (6.95,1.4) .. (10.93,3.29)   ;
\draw    (217,234) .. controls (232.68,203.6) and (246.09,221.94) .. (253.83,232.42) ;
\draw [shift={(255,234)}, rotate = 233.13] [color={rgb, 255:red, 0; green, 0; blue, 0 }  ][line width=0.75]    (10.93,-3.29) .. controls (6.95,-1.4) and (3.31,-0.3) .. (0,0) .. controls (3.31,0.3) and (6.95,1.4) .. (10.93,3.29)   ;
\draw   (320,243) .. controls (320,238.03) and (324.03,234) .. (329,234) .. controls (333.97,234) and (338,238.03) .. (338,243) .. controls (338,247.97) and (333.97,252) .. (329,252) .. controls (324.03,252) and (320,247.97) .. (320,243) -- cycle ;
\draw    (291,234) .. controls (306.68,203.6) and (320.09,221.94) .. (327.83,232.42) ;
\draw [shift={(329,234)}, rotate = 233.13] [color={rgb, 255:red, 0; green, 0; blue, 0 }  ][line width=0.75]    (10.93,-3.29) .. controls (6.95,-1.4) and (3.31,-0.3) .. (0,0) .. controls (3.31,0.3) and (6.95,1.4) .. (10.93,3.29)   ;
\draw   (358,243) .. controls (358,238.03) and (362.03,234) .. (367,234) .. controls (371.97,234) and (376,238.03) .. (376,243) .. controls (376,247.97) and (371.97,252) .. (367,252) .. controls (362.03,252) and (358,247.97) .. (358,243) -- cycle ;
\draw    (329,234) .. controls (344.68,203.6) and (358.09,221.94) .. (365.83,232.42) ;
\draw [shift={(367,234)}, rotate = 233.13] [color={rgb, 255:red, 0; green, 0; blue, 0 }  ][line width=0.75]    (10.93,-3.29) .. controls (6.95,-1.4) and (3.31,-0.3) .. (0,0) .. controls (3.31,0.3) and (6.95,1.4) .. (10.93,3.29)   ;
\draw   (396,243) .. controls (396,238.03) and (400.03,234) .. (405,234) .. controls (409.97,234) and (414,238.03) .. (414,243) .. controls (414,247.97) and (409.97,252) .. (405,252) .. controls (400.03,252) and (396,247.97) .. (396,243) -- cycle ;
\draw    (367,234) .. controls (382.68,203.6) and (396.09,221.94) .. (403.83,232.42) ;
\draw [shift={(405,234)}, rotate = 233.13] [color={rgb, 255:red, 0; green, 0; blue, 0 }  ][line width=0.75]    (10.93,-3.29) .. controls (6.95,-1.4) and (3.31,-0.3) .. (0,0) .. controls (3.31,0.3) and (6.95,1.4) .. (10.93,3.29)   ;
\draw    (356.5,146) -- (367.5,146) ;
\draw    (367.5,146) -- (367.5,151) ;
\draw    (394.5,266) -- (405.5,266) ;
\draw    (405.5,266) -- (405.5,271) ;
\draw  [dash pattern={on 0.84pt off 2.51pt}]  (264,120) -- (285.5,120) ;
\draw  [dash pattern={on 0.84pt off 2.51pt}]  (264,242) -- (285.5,242) ;

\draw (130,140) node [anchor=north west][inner sep=0.75pt]   [align=left] {I, P};
\draw (173,140) node [anchor=north west][inner sep=0.75pt]   [align=left] {P};
\draw (211,140) node [anchor=north west][inner sep=0.75pt]   [align=left] {P};
\draw (323,140) node [anchor=north west][inner sep=0.75pt]   [align=left] {P};
\draw (370,140) node [anchor=north west][inner sep=0.75pt]   [align=left] {P};
\draw (130,260) node [anchor=north west][inner sep=0.75pt]   [align=left] {I, P};
\draw (173,260) node [anchor=north west][inner sep=0.75pt]   [align=left] {P};
\draw (211,260) node [anchor=north west][inner sep=0.75pt]   [align=left] {P};
\draw (323,260) node [anchor=north west][inner sep=0.75pt]   [align=left] {P};
\draw (408,260) node [anchor=north west][inner sep=0.75pt]   [align=left] {P};
\draw (361,260) node [anchor=north west][inner sep=0.75pt]   [align=left] {P};
\draw (155,75) node [anchor=north west][inner sep=0.75pt]   [align=left] {T};
\draw (192,75) node [anchor=north west][inner sep=0.75pt]   [align=left] {T};
\draw (230,75) node [anchor=north west][inner sep=0.75pt]   [align=left] {T};
\draw (303,75) node [anchor=north west][inner sep=0.75pt]   [align=left] {T};
\draw (342,75) node [anchor=north west][inner sep=0.75pt]   [align=left] {T};
\draw (154,195) node [anchor=north west][inner sep=0.75pt]   [align=left] {T};
\draw (191,195) node [anchor=north west][inner sep=0.75pt]   [align=left] {T};
\draw (229,195) node [anchor=north west][inner sep=0.75pt]   [align=left] {T};
\draw (302,195) node [anchor=north west][inner sep=0.75pt]   [align=left] {T};
\draw (341,195) node [anchor=north west][inner sep=0.75pt]   [align=left] {T};
\draw (380,196) node [anchor=north west][inner sep=0.75pt]   [align=left] {T};
\draw (131,171) node [anchor=north west][inner sep=0.75pt]   [align=left] {\textbf{Induction-step}};
\draw (131,53) node [anchor=north west][inner sep=0.75pt]   [align=left] {\textbf{Base-case}};

\end{tikzpicture}
\caption{Base-case and induction-step \cite{induction_paper1}}
\label{formula}
\end{figure}
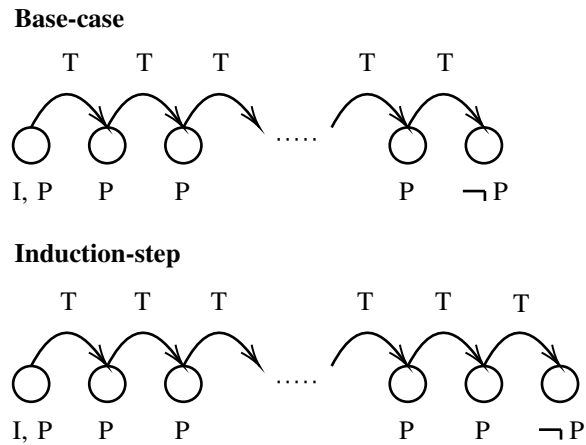

In a simple and abstract explanation to understand proof using induction, assume equations (\ref{stepa}) and (\ref{stepb}) about a design:

\begin{equation}
A_1=1
\label{stepa}
\end{equation}

\begin{equation}
\left(A_n \rightarrow A_{n+1}\right)=1
\label{stepb}
\end{equation}

Step (\ref{stepa}) proves whether the assertion holds true for the 1st clock cycle and step (\ref{stepb}) checks to prove whether the assertion $A$ holds true for a random state $A_n$ which would mean it will hold true for the next clock cycle ($A_{n+1}$) as well. If step (\ref{stepa}) and (\ref{stepb}) are true, it proves that $A$ is true for all clock cycles \cite{cadence}.

Modern \acrshort{EDA} tools such as Cadence JasperGold offer solutions like the \acrfull{SST} based on induction-based proof, where a systematic process is used to reduce the state space of a target property. The reduction is essentially done by identifying such states and writing helper assertions called lemmas. Helper assertions are properties about internal signals (states) in the design that, if proven, will be used as \say{assumptions} in the proof process of the target property (or other properties), allowing these proven properties to aid in the elimination of states \cite{cadence}. In our case, the tool pointed out a helper assertion based on the property that wasn't converging. The tool pointed out to keep the register data stable. Which helped the property to converge in a short time. Apart from all generated assertions for \acrshort{FPV}, we also used \acrfull{ABVIP} for \acrshort{SPI} protocol verification. The property set also verified the features like a daisy-chain configuration.

\singlespacing
\noindent\underline{\textit{\textbf{Aligned:}}}

While the preceding section outlines a comprehensive formal verification methodology, the totality of the verification's completeness remains uncertain. The behavioral model was implemented keeping just simulation set-up in mind. However, the prospect of creating a simulation setup solely to validate the communication between digital circuits and analog memory is deemed excessively complex for the task at hand. In terms of the \acrshort{AMS} behavioral model, a detailed disclosure of signal names is not permissible, but a generic framework can be outlined. The model encompasses input signals including 'start', 'operation', 'time\_delay\_sel', and 'analog\_start', with outputs such as 'ack' and 'ana\_start\_o'. The initiation of a transaction prompts the 'start' signal, accompanied by an operational indicator, subsequently setting a 'no\_more\_transaction' flag to prevent further transactions. The design incorporates a pipeline mechanism that induces a delay, dictated by 'time\_delay\_sel', before resetting for the next operation. Additional outputs provide external indicators like address, data, and a 'correct\_data\_flag'. The model also integrates a memory component for data storage during write operations and produces 'data\_o' for reads. The primary issue at hand is that while most inputs and outputs are digital, the few analog inputs serve only as circuit enablers, with their functionality being beyond the scope of simulation-based verification. Thus, the primary objective was to verify that the digital circuits acknowledge the enable signal from the analog circuit, without the need for a full simulation setup. Consequently, the transformation of the non-synthesizable behavioral model into a version amenable to formal verification required minimal effort and was a logical progression. The integration of this model revealed both beneficial and adverse implications, with the latter to be described subsequently: 

\begin{itemize}
\item \textbf{Incompatible Data Types:} It is a reminder that the Formal method does not accommodate analog data types. To resolve this limitation, a specialized package was integrated into the setup. This package effectively translates analog data types into digital logic, thereby circumventing the compilation errors initially encountered. As mentioned in a previous section, the evaluation of analog values has not been a priority within digital verification frameworks. This is primarily due to the inherent design of formal verification methods, which are optimized for discrete value checking rather than continuous analog values. Consequently, while the package enables the tool to process analog data types by converting them to a digital equivalent, it does not facilitate the verification of their correctness in the analog domain.

\item \textbf{Timing:} A further challenge encountered in the system setup pertained to the treatment of delays. Delays represent non-synthesizable elements that impede achieving feature equivalence. Specifically, in our configuration, handshake signals like ack signals were subject to timing delays. These delays corresponded to an approximate number of clock cycles. To address this, single-bit handshake signals were delayed using flip-flops to match the number of cycles. However, it is critical to acknowledge that the timing of these signals' arrival did not precisely mirror real-time behavior. Although the implemented delay was quantitatively aligned with clock cycles, the lack of exactness means that the solution was an approximation rather than a precise replication of real-time signal timing.

\item \textbf{Equivalence check:} The system design incorporated certain unsynthesizable constructs, notably the \say{initial begin}  and \say{task} block. It is recognized that these constructs, while commonly used in simulation environments, do not contribute to the functionality of hardware modules post-synthesis and hence can be substituted with synthesizable constructs such as \say{always} blocks. Transitioning from unsynthesizable to synthesizable constructs could potentially result in a disparity between the original and revised versions of the modules, making it imperative to ascertain their equivalence. Establishing this equivalence is essential to ensure the integrity of the design's functionality after synthesis. This aspect of verification, while critical, is identified as an area for future investigation and is beyond the scope of the current paper. However, these changes didn't prove any hinderance with core focus of the functionality check. 
\end{itemize}

The favorable aspects of the model's integration, such as the limited gate count, facilitated a seamless conversion. As highlighted in earlier discussions, this forward-looking approach not only mitigates the risk of bugs but also paves the way for the evolution of new methodologies, thereby reinforcing the robustness of formal verification at an industrial scale.

\section{Results}
\label{results}
The development took one full working week and properties were proven in minimum time. The following bugs were reported through different applications of formal verification:

\begin{itemize}
\item \textbf{Width mismatch:} Width mismatch was detected with some connections and was reported. The fix was done long before the final release
\item \textbf{Reset problem:} There was a register reset problem. As the register was not being reset on starting a new transaction, the expected data did not match the actual data.
\item \textbf{Access policy mismatch:} Some register fields were writable even though being specified as read-only
\item \textbf{Special case bugs:} There was no error response implemented for \acrshort{SPI}. For example, if the sequential read/write was interrupted in between, it not only maligned the ongoing transaction but also the next one.  
\end{itemize}

\section{Conclusion}
\label{conclusion}
The state-of-the-art formal verification methods were successfully deployed. The results were derived in a short amount of time. The primary reason to incorporate the analog behavioral model was to gain confidence over the behavioral model as that is also delivered to the end project owners along with RTL. The equivalence checking of the synthesizable model with non-synthesizable version is a future work. The inclusion of the analog behavioral model not only contributed to the completeness but also narrowed the possibility of a bug escape on scaling up the \acrshort{AMS} model. Not only was the primary goal of gaining confidence in the design successfully attained, but the process of reconfiguring the entire setup in response to specification changes was also made significantly more efficient.

\printbibliography

\end{document}